%% file: ms.tex
\documentclass[runningheads]{llncs}

\usepackage{graphicx}
\usepackage{times}
\usepackage{epsfig}
\usepackage{comment}
\usepackage{marvosym}
\usepackage{amsmath,amssymb}
\usepackage{color}
\usepackage{url}
\usepackage{booktabs}
\usepackage{makecell}
\usepackage{microtype}
\usepackage[super]{nth}
\usepackage{tabularx}
\usepackage[keeplastbox]{flushend}
\usepackage[caption=false]{subfig}
\usepackage{fancyhdr}
\usepackage{xr-hyper}
\usepackage{hyperref}
\usepackage[capitalize]{cleveref}

\usepackage{xspace}
\makeatletter
\DeclareRobustCommand\onedot{\futurelet\@let@token\@onedot}
\def\@onedot{\ifx\@let@token.\else.\null\fi\xspace}

\def\eg{\emph{e.g}\onedot} 
\def\ie{\emph{i.e}\onedot} 
\def\cf{\emph{cf}\onedot}

\def\etal{\emph{et al}\onedot}
\makeatother

\begin{document}

\title{TxT: Crossmodal End-to-End Learning with Transformers}

\author{Jan-Martin O. Steitz\inst{1} \and
Jonas Pfeiffer\inst{1} \and
Iryna~Gurevych\inst{1,2} \and
Stefan Roth\inst{1,2}}

\authorrunning{J.-M. O. Steitz et al.}

\institute{Department of Computer Science, TU Darmstadt, Germany \and
hessian.AI, Germany}

\maketitle
\thispagestyle{fancy}  % print footer on first page

\begin{abstract}
\input{tex/abstract}
\end{abstract}

\input{tex/introduction}
\input{tex/related_work}

\input{tex/method}
\input{tex/experiments}
\input{tex/conclusion}

{\small
\subsubsection{Acknowledgements.}
This work has been funded by the LOEWE initiative (Hesse, Germany) within the emergenCITY center.
}

\bibliographystyle{splncs04}
\bibliography{mainbib}

\clearpage
\input{tex/supplemental}

\end{document}

%% file: tex/abstract.tex
Reasoning over multiple modalities, \eg in Visual Question Answering (VQA), requires an alignment of semantic concepts across domains.
Despite the widespread success of end-to-end learning, today's multimodal pipelines by and large leverage pre-extracted, fixed features from object detectors, typically Faster R-CNN, as representations of the visual world. 
The obvious downside is that the visual representation is not specifically tuned to the multimodal task at hand.
At the same time, while transformer-based object detectors have gained popularity, they have not been employed in today's multimodal pipelines.
We address both shortcomings with \emph{TxT}, a transformer-based crossmodal pipeline that enables fine-tuning both language and visual components on the downstream task in a fully end-to-end manner.
We overcome existing limitations of transformer-based detectors for multimodal reasoning regarding the integration of global context and their scalability.
Our transformer-based multimodal model achieves considerable gains from end-to-end learning for multimodal question answering.  

%% file: tex/introduction.tex
\section{Introduction}

Vision and language tasks, such as Visual Question Answering (VQA) \cite{Antol:2015:VQA,goyal:2019:VQAv2,Teney:2018:TTV}, are inherently multimodal and require aligning visual perception and textual semantics.
Many recent language models are based on BERT \cite{Devlin2018}, which is based on transformers~\cite{vaswani2017attention} and pre-trained solely on text.
The most common strategy to turn such a language model into a multimodal one is to modify it to reason additionally over image embeddings, which are typically obtained through some bottom-up attention mechanism~\cite{Anderson:2018:BTA} and pre-extracted.
The consequence is that the visual features cannot be adjusted to the necessities of the multimodal reasoning task during training.
Jiang \etal \cite{Jiang:2020:IDG} recently showed that fine-tuning the visual representation together with the language model can benefit the accuracy in a VQA task. 
Pre-trained image features are thus not optimal for the multimodal setting, even when pre-training with semantically diverse datasets, e.g.~Visual Genome \cite{Krishna:2017:VGC}.
In this work, we follow \cite{Jiang:2020:IDG} to ask how the visual representation can be trained end-to-end within a multimodal reasoning pipeline.
In contrast, however, we focus on recent transformer-based architectures \cite{Chen2019,Devlin2018} with their specific requirements.

Multimodal pipelines typically \cite{Teney:2018:TTV} rely on visual features from the popular Faster R-CNN object detector \cite{Ren:2016:FRT}.
As Faster R-CNN makes heavy use of subsampling to balance positive and negative examples during training and needs non-maximum suppression to select good predictions, it is not easily amenable to end-to-end training as desired here.
Therefore, \cite{Jiang:2020:IDG} remodels Faster R-CNN into a CNN that produces dense feature maps, which are combined with an LSTM-based model \cite{Yu:2018:MFP} and trained end-to-end.
It is difficult though to combine such dense features with the latest transformer-based language models, \eg \cite{Chen2019,Li2020Oscar,lu2019vilbert,tan2019lxmert}, as the large number of image features causes scalability issues in the underlying pairwise attention mechanism.
We thus follow a different route, and instead propose to employ an alternate family of object detectors, specifically the recent Detection Transformers (DETR) \cite{Carion:2020:EOD} and variants, which treat the detection task as a set-prediction problem.
DETR produces only a small set of object detections that do not need any treatment with non-maximum suppression as its transformer decoder allows for the interaction between the detections.
We are thus able to employ the full object detector and only need to reason over a comparatively small set of image features in the BERT model.
However, a number of technical hurdles have to be overcome. 

Specifically, we make the following contributions:
\emph{(i)} We introduce alternative region features for image embeddings produced by DETR \cite{Carion:2020:EOD} and Deformable-DETR \cite{Zhu:2021:DTE}, two modern, transformer-based approaches to object detection.
Out of the box and despite their competitive accuracy for pure object detection, these transformer-based detectors deteriorate the VQA accuracy compared to the common Faster R-CNN bottom-up attention mechanism \cite{Anderson:2018:BTA}.
\emph{(ii)} We show that this accuracy loss stems from the global context of the detected objects not being sufficiently represented in the bottom-up features.
We mitigate this effect through additional global context features, which bring the VQA accuracy much closer to the baseline.
While transformer-based detectors, such as DETR, are powerful, they are also very computationally heavy to train; faster to train alternatives, \ie Deformable-DETR \cite{Zhu:2021:DTE}, are less efficient at test time, which is also undesirable.
\emph{(iii)} We address this using a more scalable variant of Deformable-DETR that still leverages multi-scale information by querying the multi-scale deformable attention module with only one selected feature map instead of a full multi-scale self-attention.
This allows retaining the training efficiency of \cite{Zhu:2021:DTE}, yet is comparably fast at test time as DETR.
\emph{(iv)} Our analysis shows that Deformable-DETR, while being a stronger object detector than DETR, also modifies the empirical distribution of the number of detected objects per image.
We find that this negatively impacts the VQA accuracy and trace this effect to the use of the focal loss \cite{Lin:2020:FLD} within Deformable-DETR.
\emph{(v)} Our final transformer-based detection pipeline enables competitive VQA accuracy when combined with transformer-based language models despite having a shorter CNN backbone compared to \cite{Ren:2016:FRT}.
More importantly, the full model including the visual features can be trained end-to-end, leading to accuracy gains of 2.3\% over pre-extracted DETR features and 1.1\% over Faster R-CNN features on the VQAv2 dataset \cite{goyal:2019:VQAv2}.

%% file: tex/related_work.tex
\section{Related Work}

With the advancement of transfer learning, the current modus operandi for multimodal learning is to combine pre-trained models of the respective modality by learning a shared multimodal space.
We thus review approaches for single modalities before turning to crossmodal reasoning.

\subsection{Learning to analyze single modalities}

\paragraph{Language understanding.}
Recently, transfer learning has dominated the domain of natural language processing, where pre-trained models achieve state-of-the-art results on the majority of language tasks. These models are predominantly trained with self-supervised objectives on large unlabeled text corpora, and are subsequently fine-tuned on a downstream task \cite{Ruder2018ULMFit,peters2017semi}. Most recent language models leverage the omnipresent transformer architectures \cite{vaswani2017attention} and are trained predominantly  with Masked-Language-Modelling (MLM) objectives as \textit{encoders} \cite{Devlin2018,Liu2019}, with next-word prediction objectives as \textit{generative/decoder} models \cite{Lewis20BART,Raffel:2020t5}, or as \textit{sequence-to-sequence} models  \cite{brown:2020,radford:2018,radford:2019}. 
 
\paragraph{Visual scene analysis.}
Convolutional neural networks (\eg, ResNet \cite{He:2016:DRL}) are widely used to encode raw image data. 
While originating in image classification tasks on datasets like ImageNet \cite{russakovsky:2015:imagenet}, they are much more broadly deployed with transfer learning as backbone for other tasks like object detection \cite{Girshick:2016:RCN} or semantic segmentation \cite{Shelhamer:2017:FCN}.

The dominant methods for object detection are Faster R-CNN \cite{Ren:2016:FRT} and variants like Feature Pyramid Networks (FPN) \cite{lin:2017:fpn} due to their high accuracy \cite{huang:2017:sat}.
Faster R-CNN has a two stage architecture with a shared backbone: 
First, a region proposal network suggests regions by a class-agnostic distinction between foreground and background.
After non-maximum suppression, the highest scoring regions are fed to the region-of-interest~(RoI) pooling layer.
RoI pooling extracts fixed-sized patches from the feature map of the shared backbone, spanning the respective regions.
These are sent to the second stage for classification and bounding box regression.
A final non-maximum suppression step is necessary to filter out overlapping predictions.
Single-stage object detectors, \eg SSD~\cite{Liu:2016:SSM} and YOLO \cite{Redmon:2016:YOL}, offer an alternative, but still need sampling of positive and negative examples during training, hand-crafted parts for \eg anchor generation, and a final non-maximum suppression step.
More recently, Detection Transformers~(DETR) \cite{Carion:2020:EOD} were proposed with a transformer-based architecture, which is conceptually much simpler and can be trained without any hand-crafted components.
Deformable-DETR~\cite{Zhu:2021:DTE} introduced a multi-scale deformable attention module to address DETR's long training time and its low accuracy for small objects.

\subsection{Learning multimodal tasks}

\paragraph{Overview.}
Most recent work on crossmodal learning relies on combining pre-trained single-modality models to learn a shared multimodal space. 
To do so, both image and text representations are passed into a transformer model \cite{Chen2019,Li2020Oscar,lu2019vilbert,tan2019lxmert}, where a multi-head attention mechanism reasons over the representations of both modalities. 
The transformer model is initialized with the weights of a pre-trained language model.
While word-embeddings represent the text input, raw images are passed through pre-trained vision models that generate encoded representations, which are passed into the transformer. 
While ResNet encodings of the entire image can be leveraged \cite{Kiela19MMBT}, it has been shown that utilizing object detection models (\ie Faster R-CNN \cite{Ren:2016:FRT}), which provide encoded representations of multiple regions of interest \cite{Anderson:2018:BTA}, benefits the downstream task \cite[\textit{inter alia}]{lu2019vilbert,tan2019lxmert}. Here, the image features are passed through an affine-transformation layer, which learns to align the visual features with the pre-trained transformer. Similarly, the pixel offsets\footnote{Depending on the model, this includes relative position, width, and height of the original image.} are used to generate positional embeddings. By combining these two representations, each object region is passed into the transformer separately.\footnote{Different approaches for combining pixel offsets with object features have been proposed, such as adding the representation (\eg, used by LXMERT \cite{tan2019lxmert}) or transforming them jointly (\eg, used by Oscar \cite{Li2020Oscar}).} 

\paragraph{Datasets.}
To learn a shared multimodal representation space, image captioning datasets such as COCO \cite{lin2014microsoft}, Flicker30k \cite{Plummer2015Flickr}, Conceptual Captions (CC) \cite{sharma-etal-2018-conceptual}, and SBU \cite{Ordonez:2011:im2text}) are commonly utilized.  The self-supervised objectives are largely the same among all approaches: next to MLM on the text part, masked feature regression, masked object detection, masked attribute detection, and cross-modality matching.

\paragraph{Transformer approaches.}
Recent multimodal models initialize the transformer parameters with BERT \cite{Devlin2018} weights and leverage the Faster R-CNN object detection model: 
LXMERT \cite{tan2019lxmert} and ViLBERT \cite{lu2019vilbert} propose a dual-stream architecture, which provides designated language and vision transformer weights. 
A joint multi-head attention component attends over both modalities at every layer. 
UNITER \cite{Chen2019} and Oscar \cite{Li2020Oscar} propose a single-stream architecture, which shares all transformer weights among both modalities.
Oscar additionally provides detected objects as input to the transformer, and argues that this allows for better multimodal grounding.
VILLA \cite{gan2020large} proposes to augment and perturb the embedding space for improved pre-training. 

\paragraph{End-to-end training.}
Above approaches combine pre-trained vision and language models, however, they do not back-propagate into the vision component. 
Thus, no capacity is given to the vision model to reason over the raw image data in terms of the downstream task; the assumption is that the pre-encoded representations are sufficient for the downstream crossmodal task. 
Kamath \etal \cite{Kamath:2021:MDETR} avoid this by incorporating multimodality into a transformer-based object detector. It is end-to-end trainable but needs computationally heavy pre-training.
Jiang \etal \cite{Jiang:2020:IDG} address this by proposing to extract the Faster R-CNN weights from \cite{Anderson:2018:BTA} into a CNN. 
They are able to leverage multimodal end-to-end training, but use a model based on long short-term memory (LSTM) \cite{Hochreiter:1997:LSTM}. The sequential processing of LSTM models hinders parallelization and thus limits the sequence length.
Pixel-BERT~\cite{Huang:2020:AIP} proposes to embed the image with a CNN and reasons over all resulting features with a multimodal transformer. 
As such their model is end-to-end trainable, but very heavy on computational resources because of the pairwise attention over these dense features.
For VL-BERT \cite{Su:2020:VLB}, a Faster R-CNN is used to pre-extract object bounding boxes. The region proposal network of Faster R-CNN is then excluded in the further training process and the language model is jointly trained with the object detector in a Fast R-CNN \cite{Girshick:2015:FRC} setting.
This separates region proposals from object classification and localization for test time, hence multi-stage inference is necessary and no full end-to-end training is possible.

To mitigate this, we propose TxT, a transformer-based architecture that combines transformer-based object detectors with transformer-based language models and can be trained end-to-end for the crossmodal reasoning task at hand.

%% file: tex/method.tex
\section{Transformers for Images and Text}
\label{sec:method}
Transformers \cite{vaswani2017attention} not only quickly developed into a standard architecture for language modeling \cite{Devlin2018}, but their benefits have recently been demonstrated also in visual scene analysis \cite{Carion:2020:EOD,Zhu:2021:DTE,Dosovitskiy:2021:ViT}.
We aim to bring these two streams of research together, which aside from conceptual advantages will allow for end-to-end learning of multimodal reasoning.

\subsection{Transformer-based object detection}
\label{sec:method-detection}
\paragraph{DETR.}
The Detection Transformer (DETR) \cite{Carion:2020:EOD} is a recent transformer-based object detector, which treats object detection as a set-prediction problem and, therefore, obviates the need for non-maximum suppression. 
It outperforms the very widely used Faster R-CNN \cite{Ren:2016:FRT} for object detection on standard datasets like COCO \cite{lin2014microsoft}. 
DETR first extracts a feature map from the input image using a standard CNN backbone. 
The features are then projected to a lower dimensionality through $1\!\times\!1$ convolutions. 
A subsequent transformer encoder uses multi-head self-attention layers and pointwise feed-forward networks to encode the features. 
These are then passed to the decoder, where a fixed set of learned object queries is used to embed object information by cross-attention with the encoder output. 
Finally, feed-forward networks predict classes and bounding boxes. 
The predictions are connected by bi-partite matching with the ground truth and a set-based loss is used for training. 
As such, DETR is completely end-to-end trainable without hand-crafted parts.

\paragraph{Deformable-DETR.} Deformable-DETR \cite{Zhu:2021:DTE} addresses two important drawbacks of the original DETR architecture: 
The long required training time (500 epochs are needed for full training) and its relatively low detection accuracy for small objects.
Deformable-DETR replaces the standard multi-head attention layers with multi-scale deformable attention modules in its transformer layers.
Instead of a pairwise interaction between all queries and keys, they only attend to a small set of sampling points around each reference point, interpolated at learned position offsets.
To incorporate multi-scale information, the attention modules work on a set of feature maps of different scales and each query feature attends across all scales. 
Additionally, in Deformable-DETR the focal loss \cite{Lin:2020:FLD} is used for matching between predictions and ground truth as well as scoring the classification. 
Deformable-DETR is not only able to reduce the total training time to only one \nth{10} of that of DETR, but even outperforms it for object detection on standard datasets \cite{lin2014microsoft}.

\begin{table*}[t]
    \centering
    \caption{\emph{Comparison of object detector architectures} on the COCO validation set \cite{lin2014microsoft} in terms of average precision (AP, higher is better) using the COCO detection evaluation metrics. 
    The results for DETR, DETR-DC5, and Faster R-CNN are quoted from \cite{Carion:2020:EOD} and the multi-scale Deformable-DETR evaluation is taken from \cite{Zhu:2021:DTE}. 
    The (+) in Faster R-CNN indicates that the models were trained with an additional GIoU loss. 
    The GFLOPs are measured over the first 100 images of the COCO validation set. 
    Deviating from \cite{Carion:2020:EOD,Zhu:2021:DTE}, which estimate the number of floating point operations and omit part of the model, we log all CUDA calls with NVIDIA Nsight Compute, therefore recording higher -- but more reliable -- numbers.}
    \label{tab:detr_comparison}
    %\vspace{-3pt}
    \begin{tabularx}{\linewidth}{@{}Xccccccccc@{}}
        \toprule
        Model & GFLOPs & \#params & Epochs & AP & AP\textsubscript{50} & AP\textsubscript{75} & AP\textsubscript{S} & AP\textsubscript{M} & AP\textsubscript{L} \\
        \midrule
        Faster R-CNN-FPN+ & 253 & 41M & 109 & 42.0 & 62.1 & 45.5 & 26.6 & 45.4 & 53.4 \\
        Faster R-CNN-R101-FPN+ & 360 & 60M & 109 & 44.0 & 63.9 & 47.8 & 27.2 & 48.1 & 56.0  \\
        \midrule
        DETR & 161 & 41M & 500 & 42.0 & 62.4 & 44.2 & 20.5 & 45.8 & 61.1  \\
        DETR-DC5 & 490 & 41M & 500 & 43.3 & 63.1 & 45.9 & 22.5 & 47.3 & 61.1 \\
        \midrule
        Deformable-DETR \textit{query stride 16} & 177 & 40M & 50 & 42.9 & 62.9 & 45.9 & 24.2 & 46.1 & 60.4 \\
        Deformable-DETR \textit{multi-scale} & 337 & 41M & 50 & 45.4 & 64.7 & 49.0 & 26.8 & 48.3 & 61.7 \\
        \bottomrule
    \end{tabularx}
\end{table*}

\subsection{Global context for DETR models}
\label{sec:method-context}
Before we can bring together transformer architectures for object detection and language modeling in an end-to-end fashion, we first need to assess whether transformer-based detectors yield a competitive baseline.
To that end, we use pre-extracted features from pre-trained detectors in visual question answering.

Surprisingly, we find that replacing Faster R-CNN features by those from DETR leads to a noticeable 2.5\% accuracy loss in terms of VQAv2 \cite{goyal:2019:VQAv2}, even if DETR is a stronger detector on standard object detection benchmarks \cite{lin2014microsoft}.
Looking at this more closely, we find that in contrast to Faster R-CNN, DETR and Deformable-DETR encode the semantic information in a single feature vector for each query in the transformer-based decoder.
We posit that their transformer encoder-decoder structure leads to learning more object-specific information in its feature vectors and global scene context not being captured.
As this global scene context is arguably important for VQA and other crossmodal tasks, we compensate for the loss in global context by \emph{augmenting each output feature vector with context information from global average pooling}.
We analyze different positions for extracting this contextual information and find a \emph{substantial gain of 1.3\% accuracy} for VQAv2 from the global context. Full details are described in \cref{sec:experiments-ablation}.
While not completely closing the accuracy gap to Faster R-CNN features, we note that these are more high-dimensional and not amenable to end-to-end crossmodal training.
Also note that adding the same contextual information to Faster R-CNN features, in contrast, does not aid VQA accuracy.
Consequently, we conclude that \emph{global contextual information about the scene is important for using transformer-based detectors for VQA}.

\subsection{Scalable Deformable-DETR}
\label{sec:method-scalable}
Another significant impediment in the deployment of transformer-based architectures for multimodal tasks are the very significant training times of DETR \cite{Carion:2020:EOD}.
This has been very recently addressed with Deformable-DETR \cite{Zhu:2021:DTE}, which is much faster to train, but is slower at test time, requiring more than double the computations compared to DETR.
We address this through a \emph{more scalable Deformable-DETR} method, which can still use the multi-scale information but is computationally more lightweight.
Deformable-DETR extracts the features maps of the last three ResNet blocks with strides of 8, 16, and 32. 
It also learns an additional convolution with stride 2 to produce a fourth feature map with stride 64. 
Those four feature maps are then fed to the multi-scale deformable attention layers of the encoder. 
In the deformable self-attention, each feature then attends to a fixed number of interpolated features of each of the feature maps, incorporating multi-scale information in the process.

Instead of performing a full self-attention, where query and keys are the same set, we use the feature maps of all scales as keys but propose to query only with the feature map of one chosen scale.
This results in \emph{a single feature map that still incorporates multi-scale information}, because the query features attend to the keys across all scales.
All further attention layers then work on this single feature map, reducing the amount of computational operations needed.
By choosing feature maps of different strides as the query, we can scale the computational complexity and accuracy corresponding to our task.
See \cref{supp:scalable-deformable} in the supplemental material for detailed results. 

\begin{figure}[t]
    \centering
    \includegraphics[width=\linewidth]{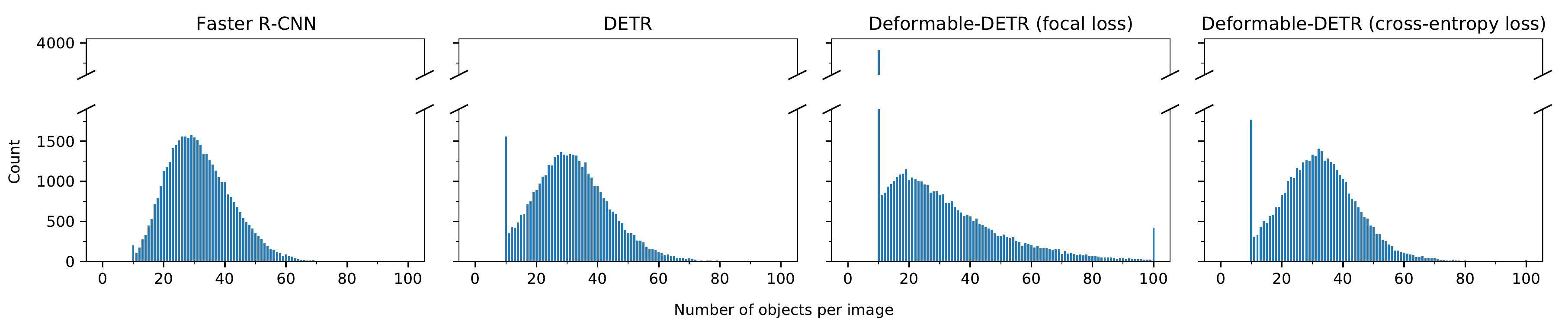}
    \vspace{-15pt}
    \caption{\emph{Number of extracted objects per image for COCO validation.} 
    Only predictions with a confidence exceeding a threshold are extracted, but a minimum of 10 and a maximum of 100 objects are selected per image. 
    Thresholds are chosen such that the total number of extracted regions is the same for all methods. 
    Deformable-DETR with focal loss produces a skewed distribution while using a cross-entropy loss resembles the detection distribution of Faster R-CNN and DETR.}
    \label{fig:region_hist}
\end{figure}

For a good trade-off between computational cost and AP to be used for our crossmodal architecture, we query with feature maps of stride 16. 
In \cref{tab:detr_comparison} we compare our model to Faster R-CNN with ResNet-50 and ResNet-101 backbones \cite{He:2016:DRL}, DETR, DETR-DC5 (with dilated convolutions in the last ResNet block), and multi-scale Deformable-DETR. 
Compared to the standard DETR model, our proposed scalable Deformable-DETR maintains a comparable number of model parameters and reaches a \emph{0.9\% higher AP} with only 10\% more computational expense while still being trained in only 50 instead of 500 epochs.
It is also faster at test time than Faster R-CNN, while being more accurate when compared with the same backbone.
Our approach thus provides a favorable trade-off as a detector basis.

\subsection{Detection losses for crossmodal tasks}
\label{sec:method-loss}
Deformable-DETR not only aims to address the lacking training efficiency of DETR, but also its limitations for hard-to-detect objects, \eg small objects or objects with rare classes.
To that end, a focal loss \cite{Lin:2020:FLD} is employed.
To assess if this impacts its use in downstream tasks, such as VQA, we analyze both the detection output as well as its use as pre-extracted feature for VQA.
To compare the suitability of the features extracted by different object detectors, we only extract the features of objects with a confidence above a certain threshold. 
A minimum of 10 objects and a maximum of 100 objects per image are extracted following Anderson \etal \cite{Anderson:2018:BTA}. 
The thresholds of DETR and Deformable-DETR are chosen such that all models produce the same total number of objects on the COCO validation set. 
When we plot the distribution of the number of objects per image across the validation dataset, shown in \cref{fig:region_hist}, we see that DETR, which is also trained with a cross-entropy loss, produces a distribution with a peak at a similar position as Faster R-CNN, although a little broader. 
Deformable-DETR trained with a focal loss to improve on difficult objects, on the other hand, has a distribution shifted to a lower number of regions and notably more spread out and skewed. 
Since small or rarely occuring objects (currently) do not play a significant role in VQA,\footnote{Verified by measuring object sizes via gradient backtracing in UNITER \cite{Chen2019} using the Grad-CAM \cite{selvaraju:2020:gradcam} approach and selecting image regions with normalized activations greater than 0.9.} this property may not necessarily benefit the crossmodal task.
Indeed, as shown in \cref{tab:vqa_results_cached}, the \emph{focal loss degrades the VQAv2 performance}. 
Since we are less interested in pure detection accuracy, but rather in crossmodal performance, we therefore train our Deformable-DETR with a \emph{cross-entropy loss for the class labels}.
As can be seen in \cref{fig:region_hist} \emph{(right)}, this makes the detection statistics more closely resemble those of Faster R-CNN and DETR, and leads to a \emph{clear increase of 0.6\% on VQAv2 test-dev} (\cf \cref{tab:vqa_results_cached}).

\begin{figure}[t!]
    \centering
    \includegraphics[width=\columnwidth]{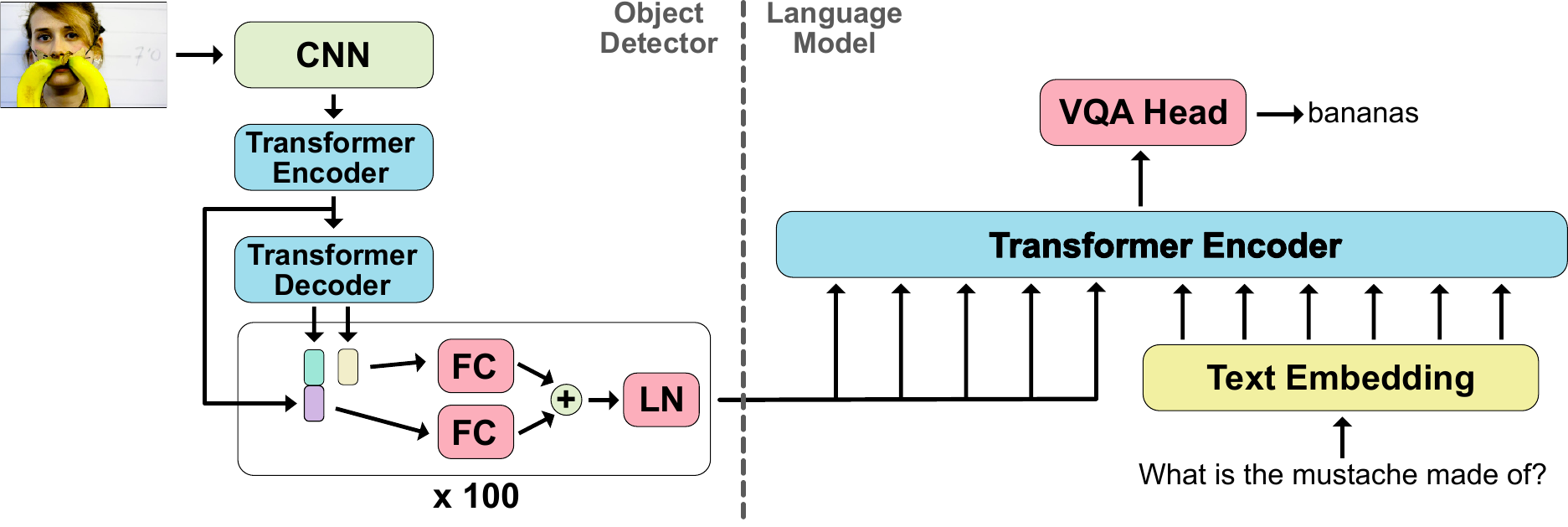}
    \vspace{-17pt}
    \caption{\emph{Overview of our proposed TxT model:} 
    A transformer-based object detector \cite{Carion:2020:EOD,Zhu:2021:DTE} produces feature representations and bounding box coordinates for its predictions \emph{(left)}. 
    Following \cite{Chen2019}, each feature and position pair is combined by fully connected (FC) layers. The aggregated representations are passed through a layer norm (LN) and utilized as object embeddings; concatenated with the sequence of text embeddings, the multimodal instance is passed into the transformer-based language model \emph{(right)}.
    A task specific head (here for VQA) predicts the answer. 
    The model is end-to-end trainable, including both the vision and language components.}
    \label{fig:txt-model}
\end{figure}

\subsection{The TxT model}
\label{sec:method-txt}
With the proposed \emph{TxT model}, we now combine transformer-based object detection (T) and transformer-based language models (T) in a crossmodal approach (x). 
It enables us to bridge the information gap between language and vision models and perform crossmodal end-to-end training while avoiding computationally heavy dense features~\cite{Huang:2020:AIP}.
We base the language model on BERT-base, pre-trained with MLM on text corpora only \cite{Devlin2018}. 
For the object detector, we employ either DETR or Deformable-DETR models as described in \cref{sec:method-detection,sec:method-context,sec:method-scalable,sec:method-loss}.
The structure of TxT is shown in \cref{fig:txt-model}.
In the object detector a CNN backbone (ResNet-50 \cite{He:2016:DRL}) computes feature representations from the input image. 
They are passed to the transformer encoder, where they interact in its multi-head attention layers. 
In the decoder, the encoded feature map gets queried for objects and for each object the feature representation and the bounding box coordinates are extracted \cite{Carion:2020:EOD} (see \cref{sec:method-scalable} for details). 
The global average-pooled encoder output is concatenated to the feature representation of each object to provide global context (\cf~\cref{sec:method-context}).
We use the complete set of predictions generated by the object detector for TxT. 
Following~\cite{Chen2019}, features and bounding box coordinates are projected by fully connected layers to the dimension of the BERT model and combined. 
After passing a normalization layer, they are used as object embeddings in the language model. 

The text input is run through a tokenizer and both embeddings, visual and textual, are concatenated to a sequence so that the self-attention of the transformer encoder is able to reason over both modalities. 
The encoder output is passed through a task-specific head, in our case a VQA head, consisting of a small multi-layer perceptron (MLP) to produce the answer predictions, again following \cite{Chen2019}.
The TxT model is trained fully end-to-end with a binary cross-entropy loss across the border between the vision and the language part. 
Thus, we are able to fine-tune the object detector specifically for the crossmodal task.

%% file: tex/experiments.tex
\section{Experiments}
\label{sec:experiments}
To evaluate our TxT model, we perform experiments on a visual questioning answering task.
Specifically, we use the VQAv2 training set \cite{goyal:2019:VQAv2} and add additional question and answer pairs from Visual Genome \cite{Krishna:2017:VGC} for augmentation, following \cite{Anderson:2018:BTA,Chen2019}, but omit using the validation split for training.
We follow standard practice and only classify for the 3129 most frequent answers \cite{Anderson:2018:BTA,Chen2019,Yu:2019:DMC}.
All results are evaluated with the standard VQA accuracy metrics \cite{Antol:2015:VQA} on the test-dev set from the 2021 challenge for VQAv2 \cite{goyal:2019:VQAv2}.

\subsection{Technical details}
\label{sec:experiments-technical}
\paragraph{Object detector.}
The transformer-based object detectors in our TxT model employ a ResNet-50 \cite{He:2016:DRL} backbone. 
The extracted features are projected to a hidden dimension of 256 of the transformer layers and then passed to an encoder consisting of 6 transformer layers. 
The transformer layers consist of multi-head attention layers with 8 heads and feed-forward networks with a dimension of 2048 for DETR and 1024 for Deformable-DETR, as described in \cite{vaswani2017attention}. 
Learned object queries generate 100 object features with an alternation of self-attention and cross-attention in the 6 transformer layers of the decoder. 
A linear projection is used to classify the object and a 3-layer perceptron is used to generate bounding box coordinates \cite{Carion:2020:EOD}. 
Deformable-DETR uses 300 queries in its default configuration. 
For end-to-end learning in TxT, we use a variant with 100 object queries. 
Deformable-DETR uses multi-scale deformable attention in its transformer layers, which employs 4 keys interpolated around each reference point of the query. 
We use Deformable-DETR with iterative bounding box refinement \cite{Zhu:2021:DTE}. Both DETR and Deformable-DETR use global context as discussed in \cref{sec:method-context}.

\paragraph{Pre-training.}
Following Anderson \etal \cite{Anderson:2018:BTA}, we pre-train all object detectors on the Visual Genome dataset \cite{Krishna:2017:VGC} with a maximum input image size of 600$\times$1000 and add an additional head for the prediction of object attributes.
We train DETR for 300 epochs on 8 NVIDIA V100 GPUs with 4 images per GPU. 
We use a learning rate of 1e-4 for DETR and a learning rate of 1e-5 for the ResNet backbone. 
After 200 epochs, we drop the learning rate by a factor of 10. 
For all other settings, we follow \cite{Carion:2020:EOD}.
The training protocol for Deformable-DETR follows \cite{Zhu:2021:DTE}.

\paragraph{Language model.}
The language model of TxT is based on BERT-base \cite{Devlin2018} with 12 transformer layers and 768 hidden dimensions. BERT is pre-trained on the combination of the BooksCorpus \cite{Zhu2015BooksCorpus} and the entire English Wikipedia corpus, consisting of 3.3 billion words in total.
A WordPiece tokenizer \cite{SchusterN12WordPiece} is trained on the entire corpus, amounting to a vocabulary size of 30k, with each token corresponding to a position in the designated word-embedding matrix.
\textit{Positional} embeddings are trained  to preserve the sentence order, and \textit{segment} embeddings are trained to indicate if a token belongs to sentence A or B. The embedding representation,  which is passed into the transformer, thus amounts to the sum over \textit{word}-, \textit{positional}-, and \textit{segment}-embedding of each respective token. 
BERT is pre-trained with an MLM,\footnote{Input tokens are randomly masked, with the objective of the model being to predict the missing token given the surrounding context.} as well as a next-sentence prediction objective.\footnote{Two sentences A and B are passed into the model, where 50\% of the time B follows A in the corpus, and 50\% of the time B is randomly sampled. The objective is to predict whether or not B is a negative sample.}

\paragraph{VQA training.}
The TxT model is trained for 10k iterations on 4 GPUs and a learning rate of 8e-5. 
We employ a binary cross-entropy loss and the AdamW \cite{loshchilov:2019:adamw} optimizer.
We use a linear warmup of 600 iterations and a linear decay of the learning rate afterwards. 
In case of pre-extracted, fixed object features, we train the TxT model similar to \cite{Chen2019} with 5120 input tokens and for each iteration we accumulate the gradients over 5 steps.
For end-to-end training, we pre-train the TxT model with fixed features for 4k iterations for DETR and 6k iterations for Deformable-DETR with the above settings.
Then, we train the model fully end-to-end for 10k iterations with 8 images and questions per batch and accumulate the gradients for 32 steps for each iteration to maintain the effective batch size of the fixed-feature setting.
When we employ TxT with DETR as the object detector, we initialize the learning rate for the DETR and CNN backbone components  with 3e-4 and  3e-5, respectively. 
Deformable-DETR is trained with a learning rate of 6e-4 and 6e-5 for the backbone.
When TxT is used with DETR and thresholding the predictions, we apply a factor of 10 to the learning rate of the class-prediction layer of DETR. For end-to-end training of TxT, the learning rates and training schedule were determined with a coarse grid search over the hyperparameters.

\subsection{Representational power of visual features}
\label{sec:experiments-richness}
In a first step, we investigate if DETR and Deformable-DETR are able to produce object feature descriptors whose representational power for VQA tasks is comparable to the widely used Faster R-CNN;
we assess this through the VQA accuracy.
We pre-extract object features and bounding box coordinates with all object detectors.
Following standard practice (\eg \cite{Anderson:2018:BTA,Chen2019,Su:2020:VLB}), all objects with a class confidence above a certain threshold are extracted, but a minimum of 10 and a maximum of 100 objects are used per image. 
We calibrate the confidence threshold of DETR and Deformable-DETR on COCO validation so that the total number of extracted object features is 1.3 million, comparable to Faster R-CNN.
The TxT model is then trained with the pre-extracted features.
When used with Faster R-CNN features, this is equivalent to the UNITER-base model \cite{Chen2019} with only MLM pre-training, \ie initalized with BERT weights. 
We note that the numbers differ slightly from those in \cite{Chen2019}, because we omit the validation set when training for test-dev submission and use the most recent VQA challenge 2021.

The results in \cref{tab:vqa_results_cached} show that DETR and Deformable-DETR are able to produce object features with a representational power that \emph{competitively supports VQA tasks}, especially considering their low dimensionality of 256 (compared to 2048 dimensions of Faster R-CNN features).
Also, these results are achieved with only a ResNet-50 backbone (compared to ResNet-101 for Faster R-CNN).
While DETR leads to a $\sim\kern-0.25em 1.2$\% loss in VQA accuracy compared to Faster R-CNN, it is more efficient at test time (see \cref{tab:detr_comparison}) and, more importantly, allows for end-to-end training (\cf \cref{sec:experiments-e2e}).

Moreover, we find that Deformable-DETR (with query stride 16) as the object detector for VQA performs less well than DETR, about 0.5\% worse in VQA accuracy.
As analyzed in \cref{sec:method-loss}, this is traceable to the focal loss.
We find that a cross-entropy loss is a more suitable pre-training objective for Deformable-DETR as object detector for the VQA task, leading to $\sim\kern-0.25em 0.6$\% higher accuracy on VQA while maintaining the same total number of predictions.
With these modifications, \emph{Deformable-DETR performs comparatively with DETR, yet is still much faster to train}.

\begin{table}[t]
    \centering
    \caption{\emph{Comparison of VQA accuracy on pre-extracted features from different object detectors} evaluated on VQAv2 test-dev \cite{goyal:2019:VQAv2} (higher is better). 
    Deformable-DETR refers to our variant, where we query the attention module with a feature map of  stride 16 (see \cref{sec:method-scalable} for details). The best results in each column are bold, the \nth{2} best underlined.}
    \label{tab:vqa_results_cached}
    %\vspace{-3pt}
    \begin{tabularx}{\linewidth}{@{}X@{}c@{\ \ \;}c@{\ \ \;}c@{\ \ \;}c@{}}
        \toprule
        Method & number & yes/no & other & overall \\
        \midrule
        Faster R-CNN & \bfseries 51.18 & \bfseries 85.04 & \bfseries 59.00 & \bfseries 68.84 \\
        DETR & \underline{49.49} & 83.88 & \underline{57.81} & 67.60 \\
        Deformable-DETR (focal loss) & 47.59 & 83.87 & 57.18 & 67.09 \\
        Deformable-DETR (cross-entropy loss) & 48.66 & \underline{84.46} & 57.62 & \underline{67.66} \\
        \bottomrule
    \end{tabularx}
\end{table}

\subsection{Multimodal end-to-end training}
\label{sec:experiments-e2e}
As a preliminary experiment, referred to as TxT-DETR (thresholded), we only pass objects to the multimodal transformer layers where the class confidence exceeds a threshold of 0.5.
To accomplish this, we generate a mask from thresholding the class confidence. 
We then multiply the object features with the mask, thus allowing the gradient to backpropagate to the class prediction layer of DETR.
Only features and related bounding box coordinates according to the mask are then passed from the DETR part of TxT to its language model.
At the beginning of training, DETR selects around 30 objects per image, equivalent to the setting of pre-extracted features. 
During training, the TxT model learns to select more objects, saturating at around 80 per image. 
This suggests that \emph{more objects are beneficial for solving the VQA task}, which confirms an empirical observation of Jiang \etal \cite{Jiang:2020:IDG}, who showed that the accuracy starts to saturate  between 100 and 200 regions.
As \cref{tab:vqa_results} shows, including \emph{all} predicted objects in the multimodal reasoning (DETR \emph{all predictions}) without end-to-end training indeed leads to a $\sim\kern-0.25em 0.5$\% higher accuracy than TxT-DETR (thresholded), which leverages the end-to-end training.

For multimodal end-to-end training of our TxT model, we therefore employ all 100 object predictions, eliminating the need to threshold and enabling a gradient for object predictions that would be discarded otherwise.
In order to alleviate the computational and memory cost of multimodal self-attention, we pre-train Deformable-DETR with only 100 object queries (instead of the standard 300), which reduces the VQA accuracy by $\sim\kern-0.25em 0.7$\% when evaluating in the pre-extracted features setup.
To make the effect of the higher object number distinguishable from the end-to-end benefit, we also show results for pre-extracted features training with all object predictions in \cref{tab:vqa_results}.
With DETR as object detector, the gain from using all predictions is $\sim\kern-0.25em 2$\% and for Deformable-DETR it is less pronounced with a benefit of $\sim\kern-0.25em 0.5$\%.

Finally, we assess our full end-to-end trainable model.
We find that \emph{multimodal end-to-end training substantially improves the accuracy of TxT}: The TxT model with Deformable-DETR improves by $\sim\kern-0.25em 2.1$\% from 66.99\% for pre-extracted Deformable-DETR features to 69.06\% accuracy on VQAv2 test-dev.
The TxT model in combination with DETR achieves an overall gain in accuracy of $\sim\kern-0.25em 2.3$\% from 67.60\% accuracy for pre-extracted DETR features to 69.93\% accuracy for multimodal end-to-end training, thus improving by $\sim\kern-0.25em 1.1$\% over the Faster R-CNN features.
While our results are competitive with current models on an equal footing, they could be improved further with pre-training tasks like image-text matching and word-region alignment \cite{Chen2019} at the expense of significant computational overhead.

\begin{table}[t]
    \centering
    \caption{\emph{Results of the multimodal end-to-end trainable TxT model} on VQAv2 test-dev \cite{goyal:2019:VQAv2} (higher is better). 
    \emph{e2e} denotes end-to-end training.
    In the upper part of the table, we show the results of counterparts with pre-extracted features and Faster R-CNN features. The best results in each column are bold, the \nth{2} best underlined.}
    \label{tab:vqa_results}
    %\vspace{-3pt}
    \begin{tabularx}{\linewidth}{@{}X@{}c@{\ \ \;}c@{\ \ \;}c@{\ \ \;}c@{\ \ \;}c@{}}
        \toprule
        Method & e2e & number & yes/no & other & overall \\
        \midrule
        Faster R-CNN & & 51.18 & 85.04 & 59.00 & 68.84 \\
        DETR & & 49.49 & 83.88 & 57.81 & 67.60 \\
        DETR (all predictions) & & \bfseries 52.33 & 85.57 & \underline{59.84} & \underline{69.59} \\
        Deformable-DETR (100 queries) & & 47.19 & 84.05 & 56.92 & 66.99 \\
        Deformable-DETR (100 q., all predictions) & & 48.54 & 83.79 & 57.82 & 67.47 \\
        \midrule
        TxT-DETR (thresholded) & $\checkmark$ & 50.59 & 85.66 & 59.23 & 69.14 \\
        TxT-DETR & $\checkmark$ & \underline{51.73} & \bfseries 86.33 & \bfseries 60.04 & \bfseries 69.93 \\
        TxT-Deformable-DETR & $\checkmark$ & 49.70 & \underline{85.76} & 59.19 & 69.06 \\
        \bottomrule
    \end{tabularx}
\end{table}

\subsection{Ablation study}
\label{sec:experiments-ablation}
As discussed in \cref{sec:method-context}, adding global context is helping DETR \cite{Carion:2020:EOD} to produce visual features that are better suited for VQA.
We now investigate the suitable position from which to source this global information. 
We consider three positions at which to apply a global average pooling of the feature map in the DETR network: 
\emph{(i)} pooling the feature map produced by the CNN backbone, 
\emph{(ii)} after the backbone feature map is projected to the lower dimensions of the transformer layers, and 
\emph{(iii)} pooling the encoder output. 
As we can see in \cref{tab:detr_context_ablation}, \emph{all variants with added global context lead to a gain in accuracy}. 
We obtain the best results when using the \emph{encoder output}, leading to a gain of $\sim\kern-0.25em 1.3$\% in terms of VQAv2 \cite{goyal:2019:VQAv2} accuracy over using no global context. 
Pooling the feature map produced by the CNN backbone gives a slightly lower gain of $\sim\kern-0.25em 0.9$\% and requires 2304-dimensional features compared to only 512 dimensions for the encoder output. 
Projecting the backbone feature map first to the transformer dimension also gives 512-dimensional features, but yields only a $\sim\kern-0.25em 0.8$\% gain.

To ensure a fair comparison, we also verify if Faster R-CNN \cite{Ren:2016:FRT} features similarly benefit from global context.
To that end, we concatenate the Faster R-CNN features with global average-pooled features of its shared backbone. 
In contrast to DETR, this does not lead to better VQA results, rather to a slight degradation in accuracy of $\sim\kern-0.25em 0.7$\%.
We conclude that global context is important for VQA, yet is sufficiently represented in the common Faster R-CNN features.
Transformer-based detectors, in contrast, benefit significantly from the added context.

\begin{table}[t]
    \caption{\emph{Global context features} for DETR \cite{Carion:2020:EOD} and Faster R-CNN \cite{Ren:2016:FRT} from different locations in the network. 
    The results on a VQA task are evaluated on VQAv2 \cite{goyal:2019:VQAv2} test-dev (in \%, higher is better). 
    We also report the total dimensionality $d$ of the extracted features.}
    \label{tab:detr_context_ablation}
    \centering
    %\vspace{-3pt}
    \begin{tabularx}{\linewidth}{@{}X@{}c@{\ \;}c@{\ \;}c@{\ \;}c@{\ \;}c@{}}
        \toprule
        Model & $d$ & number & yes/no & other & overall \\
        \midrule
        DETR (no context) & 256 & 47.65 & 82.67 & 56.61 & 66.33 \\
        DETR (backbone context) & 2304 & 48.80 & 83.68 & 57.38 & 67.24 \\
        DETR (projected backbone context) & 512 & 48.69 & 83.43 & 57.36 & 67.12 \\
        DETR (encoder context) & 512 & 49.49 & 83.88 & 57.81 & 67.60 \\
        \midrule
        Faster R-CNN (no context) & 2048 & 51.18 & 85.04 & 59.00 & 68.84 \\
        Faster R-CNN (backbone context) & 3072 & 49.86 & 84.29 & 58.37 & 68.08 \\
        \bottomrule
    \end{tabularx}
    %\vspace{-2pt}
\end{table}

%% file: tex/conclusion.tex
\section{Conclusion}
In this paper we proposed TxT, an end-to-end trainable transformer-based architecture for crossmodal reasoning.
Starting from the observation that transformer-based architectures have not only been successfully used for language modeling, but more recently also for object detection, we investigated the use of Detection Transformers and variants as source of visual features for visual question answering.
We found that global context information as well as an adaptation of the loss function are needed to yield rich visual evidence for crossmodal tasks.
We also proposed a speed-up mechanism for multi-scale attention.
Our final end-to-end trainable architecture yields a clear improvement in terms of VQA accuracy over standard pre-trained visual features and paves the way for a tighter integration of visual and textual reasoning with a unified network architecture.

%% file: tex/supplemental.tex
\appendix

\section{Scalable Deformable-DETR variants}
\label{supp:scalable-deformable}

In \cref{sec:method-scalable} of the main paper, we propose a scalable variant of Deformable-DETR~\cite{Zhu:2021:DTE}.
There, we extract a set of multi-scale feature maps from the CNN backbone. These are passed to the deformable attention layers of the encoder.
By choosing feature maps of different strides as query in the first deformable attention layer, we can choose a suitable trade-off between computational cost and detection accuracy for the task at hand.
All following layers will then work on the single-scale result of the first encoder layer that embedded the multi-scale information during the attention.
\cref{tab:detr_comparison_small} shows that by going from coarse (stride 32) to fine feature maps (stride 8), it is possible to \emph{scale the average precision (AP) and the computational cost} evenly between single-scale and multi-scale Deformable-DETR without complex architectural changes.

\begin{table}[t]
    \centering
    \caption{\emph{Deformable-DETR variants.} We train reduced models with a ResNet-18 backbone, 3 encoder layers, input images with a maximum size of 600$\times$1000 on COCO, and evaluate the average precision (AP, in \%, higher is better) on the validation set. The GFLOPs and number of parameters are measured on full models with a ResNet-50 backbone, 6 encoder layers, and maximum image sizes of 800$\times$1333 as described in \cref{tab:detr_comparison} of the main paper.}
    %\vspace{-3pt} % for consistency
    \begin{tabularx}{\linewidth}{@{}Xc@{\ \ }c@{\ \ }c@{\ \ }c@{\ \ }c@{\ \ }c@{\ \ }c@{\ \ }c@{}}
        \toprule
        Model & GFLOPs & \#params & AP & AP\textsubscript{50} & AP\textsubscript{75} & AP\textsubscript{S} & AP\textsubscript{M} & AP\textsubscript{L} \\
         \midrule
        Single-scale & 141 & 35M & 30.9 & 51.1 & 31.3 & 10.5 & 32.7 & 51.4  \\
        Query stride 32 & 152 & 40M & 34.2 & 54.1 & 35.5 & 14.1 & 36.9 & 53.2 \\
        Query stride 16 & 177 & 40M & 36.8 & 56.1 & 38.9 & 17.4 & 39.4 & 51.1 \\
        Query stride 8 & 274 & 40M & 38.5 & 57.2 & 41.3 & 19.9 & 41.2 & 54.6 \\
        Multi-scale & 337 & 41M & 39.4 & 58.3 & 42.2 & 20.1 & 41.8 & 56.1 \\
         \bottomrule
    \end{tabularx}
    \label{tab:detr_comparison_small}
\end{table}

\section{Visualizations}

To analyze the qualitative differences between feeding the language model with pre-extracted Faster R-CNN \mbox{features /} bounding boxes and our end-to-end trained TxT-DETR model, we show side-by-side comparisons of the image regions the language model attends to.
We determine the strength the language model attends to each image region using the Grad-CAM approach \cite{selvaraju:2020:gradcam} and tap the gradients and activations of the linear layer projecting the image features to the dimensions of the language model. The importance of each region is shown by the opacity in the images.

\begin{figure}
    \centering
    \subfloat{
        \centering
        \includegraphics[width=.5\linewidth,trim=40 20 40 40, clip]{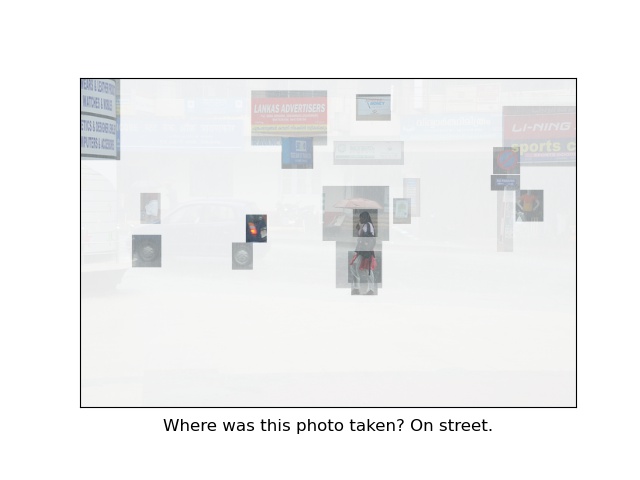}
    }%
    \subfloat{
        \centering
        \includegraphics[width=.5\linewidth,trim=40 20 40 40, clip]{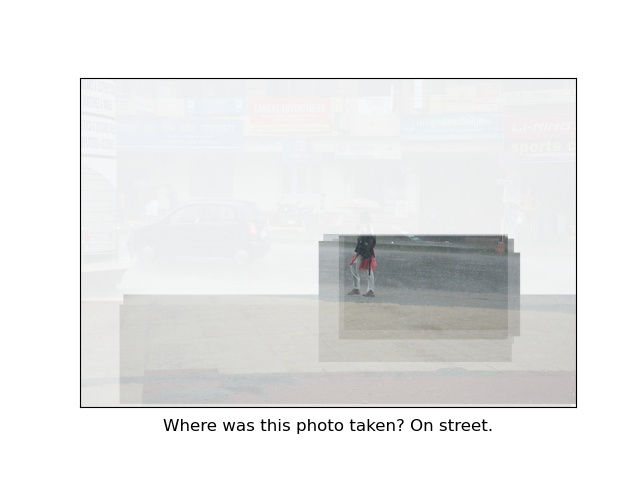}
    }%

    \subfloat{
        \centering
        \includegraphics[width=.5\linewidth,trim=40 20 40 40, clip]{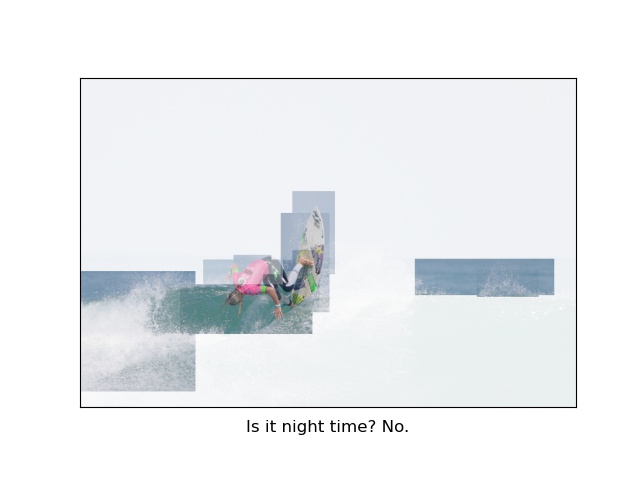}
    }%
    \subfloat{
        \centering
        \includegraphics[width=.5\linewidth,trim=40 20 40 40, clip]{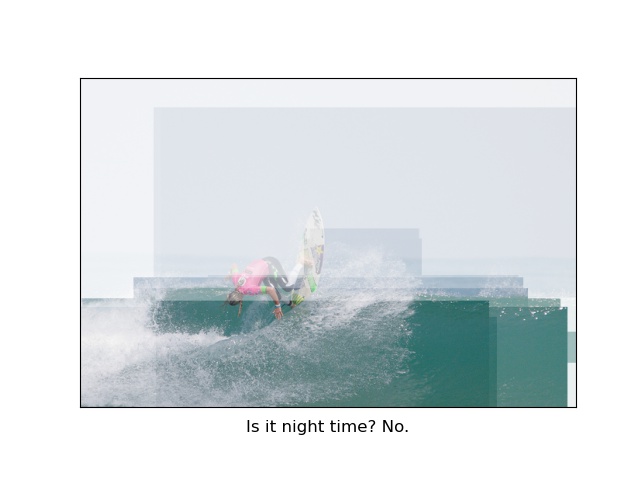}
    }%

    \subfloat{
        \centering
        \includegraphics[width=.5\linewidth,trim=40 20 40 40, clip]{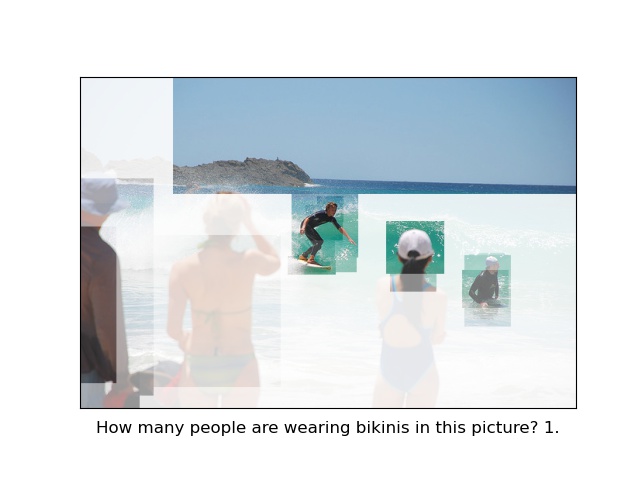}
    }%
    \subfloat{
        \centering
        \includegraphics[width=.5\linewidth,trim=40 20 40 40, clip]{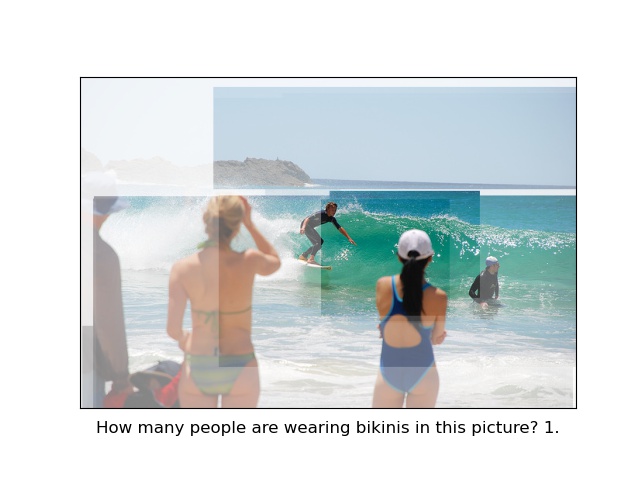}
    }%
    \caption{Comparison between the size of the image regions the language model attends to. Faster R-CNN regions are shown in the left column and TxT-DETR regions on the right.}
    \label{fig:vis}
\end{figure}

As can be seen in \cref{fig:vis}, the regions identified by Faster R-CNN are smaller and more selective to details in the images. The regions our end-to-end trained TxT-DETR model learns to produce to aid the VQA task are larger and cover bigger parts of the input images, providing a more comprehensive view of the entire scene.

\begin{figure}
    \centering
    \subfloat{
        \centering
        \includegraphics[width=.5\linewidth,trim=40 10 40 40, clip]{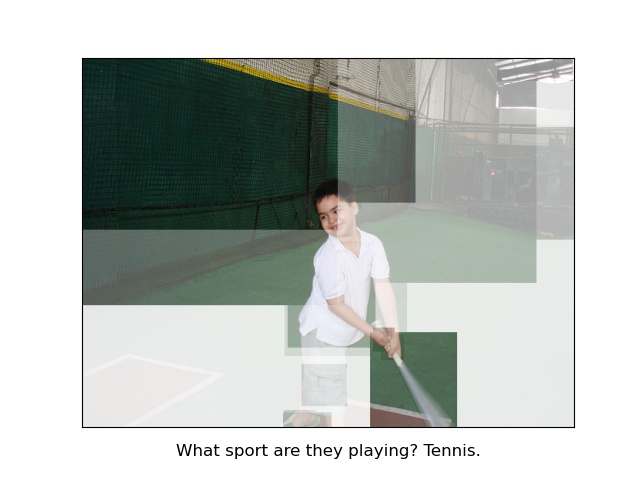}
    }%
    \subfloat{
        \centering
        \includegraphics[width=.5\linewidth,trim=40 10 40 40, clip]{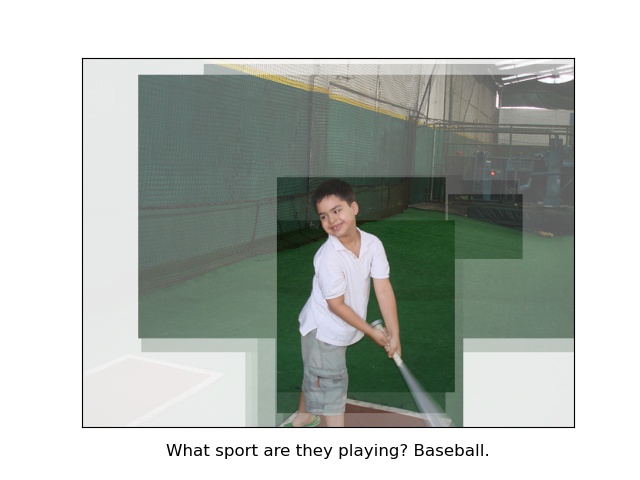}
    }%
    
    \subfloat{
        \centering
        \includegraphics[width=.5\linewidth,trim=40 10 40 40, clip]{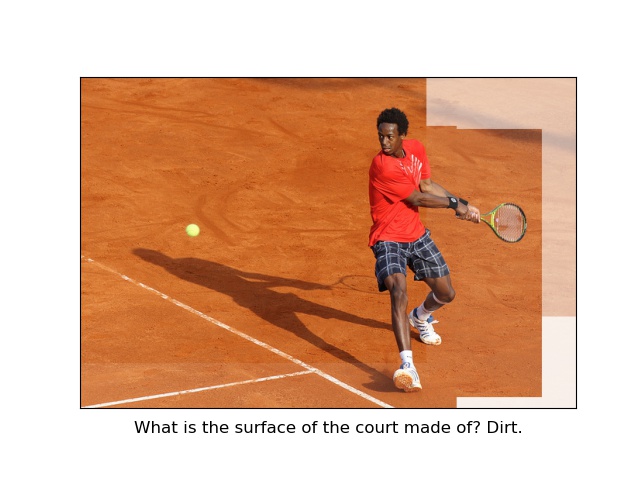}
    }%
    \subfloat{
        \centering
        \includegraphics[width=.5\linewidth,trim=40 10 40 40, clip]{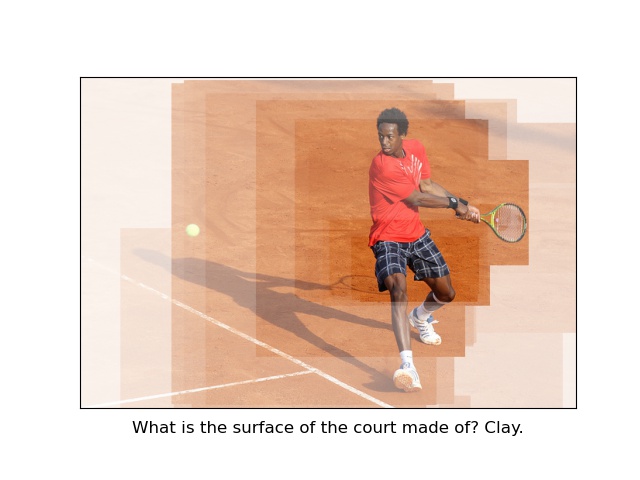}
    }%
    
    \subfloat{
        \centering
        \includegraphics[width=.5\linewidth,trim=40 10 40 40, clip]{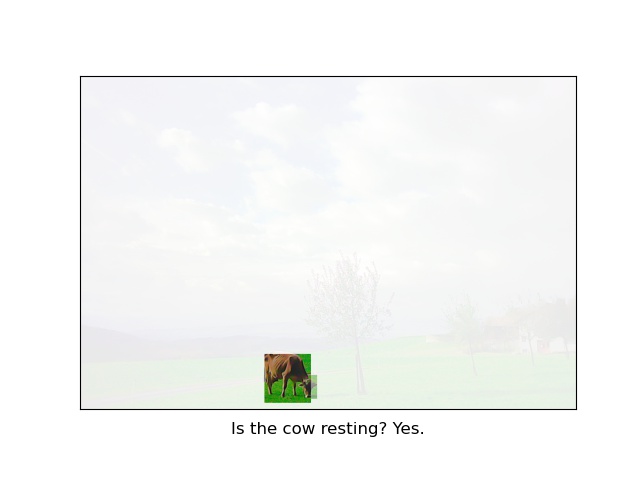}
    }%
    \subfloat{
        \centering
        \includegraphics[width=.5\linewidth,trim=40 10 40 40, clip]{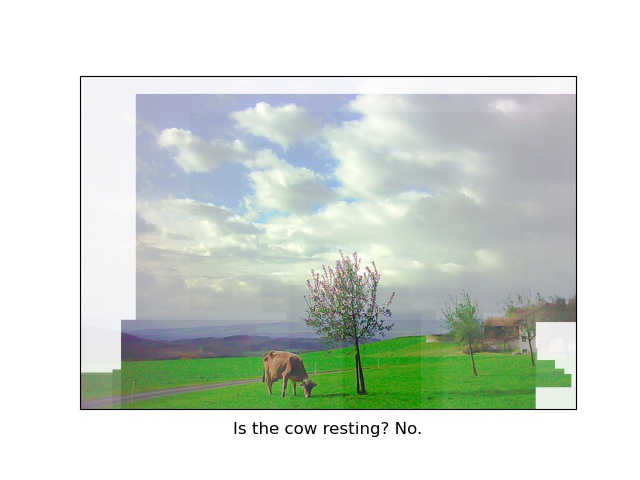}
    }%
    \caption{Examples for failure cases when the language model uses Faster R-CNN features, which TxT-DETR predicts correctly. Faster R-CNN regions are shown in the left column and TxT-DETR regions on the right.}
    \label{fig:fail_cases}
\end{figure}

\begin{figure}
    \centering
    \subfloat{
        \centering
        \includegraphics[width=.5\linewidth,trim=40 10 40 40, clip]{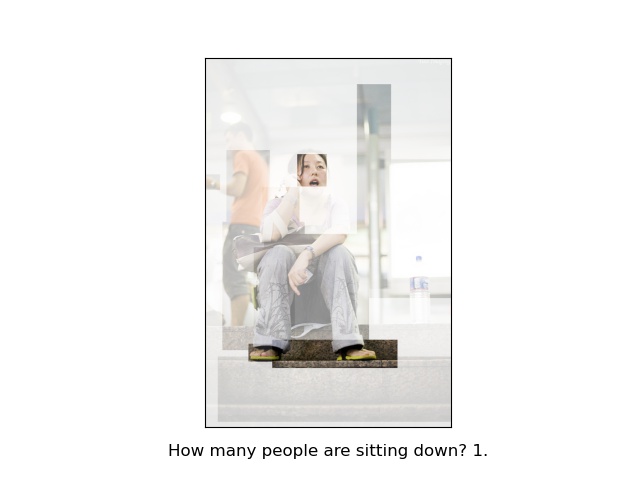}
    }%
    \subfloat{
        \centering
        \includegraphics[width=.5\linewidth,trim=40 10 40 40, clip]{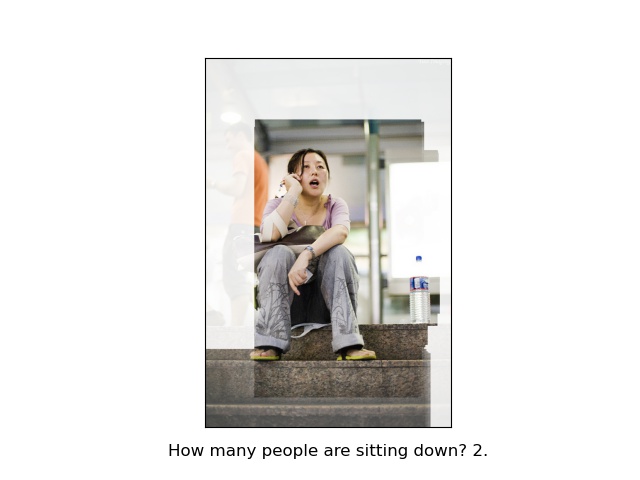}
    }%
    
    \subfloat{
        \centering
        \includegraphics[width=.5\linewidth,trim=40 10 40 40, clip]{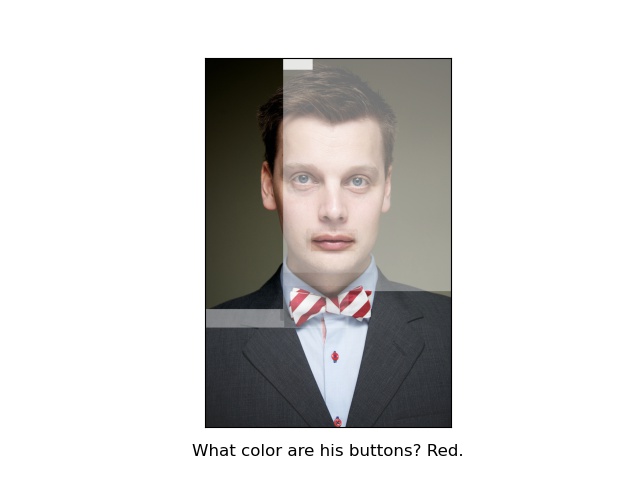}
    }%
    \subfloat{
        \centering
        \includegraphics[width=.5\linewidth,trim=40 10 40 40, clip]{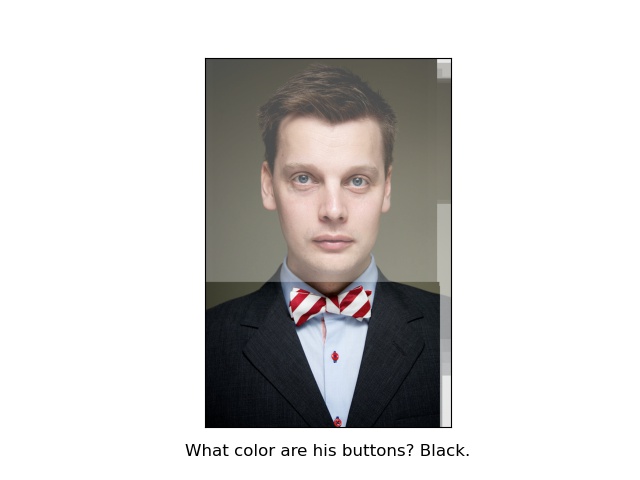}
    }%
    
    \subfloat{
        \centering
        \includegraphics[width=.5\linewidth,trim=40 10 40 40, clip]{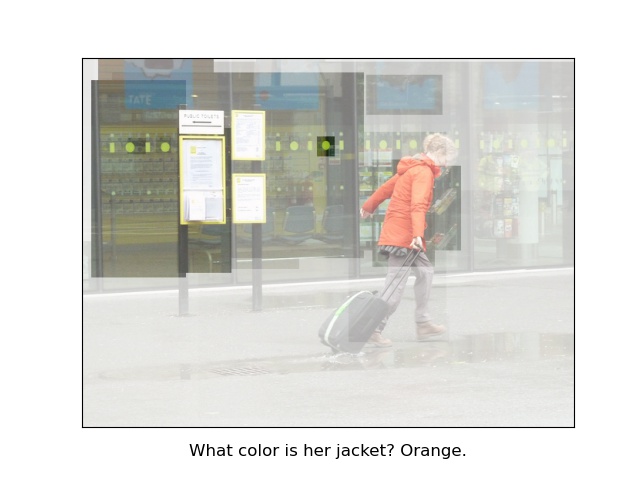}
    }%
    \subfloat{
        \centering
        \includegraphics[width=.5\linewidth,trim=40 10 40 40, clip]{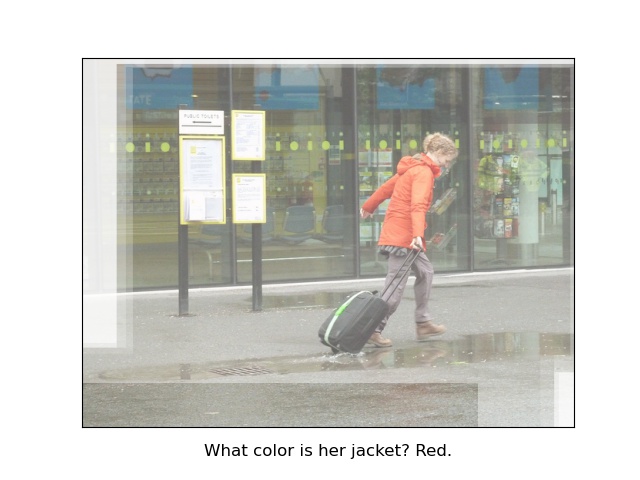}
    }%
    \caption{Examples for failure cases of TxT-DETR that the language model using Faster R-CNN features predicts correctly. Faster R-CNN regions are shown in the left column and TxT-DETR regions on the right.}
    \label{fig:fail_cases_detr}
\end{figure}

In \cref{fig:fail_cases} we show cases where the language model in combination with Faster R-CNN features fails to predict the right answer, but our end-to-end trained TxT-DETR model can predict the answer successfully; in \cref{fig:fail_cases_detr} we show failure cases for TxT-DETR, respectively.
We observe that TxT-DETR tends to focus more on the relevant objects with their context (\cref{fig:fail_cases}), though the generally more expansive capturing of the context can also hurt in certain cases (\cref{fig:fail_cases_detr}). 
The improvement through more context in terms of bounding boxes covering larger areas is akin to and consistent with the benefit of global image context encoded additionally in the visual feature vector (see \cref{sec:method-context} of the main paper).

\clearpage

